\documentclass{article}
\usepackage{spconf,amsmath,graphicx}

\title{MULTIMODAL END-TO-END GROUP EMOTION RECOGNITION USING CROSS-MODAL ATTENTION}
\name{Lev Evtodienko}
\address{Higher School of Economics}
\begin{document}
\maketitle

\begin{abstract}
Classifying group-level emotions is a challenging task due to complexity of video, in which not only visual, but also audio information should be taken into consideration. Existing works on multimodal emotion recognition are using bulky approach, where pretrained neural networks are used as a feature extractors and then extracted features are being fused. However, this approach does not consider attributes of multimodal data and feature extractors cannot be fine-tuned for specific task which can be disadvantageous for overall model accuracy. To this end, our impact is twofold: (i) we train model end-to-end, which allows early layers of neural network to be adapted with taking into account later, fusion layers, of two modalities; (ii) all layers of our model was fine-tuned for downstream task of emotion recognition, so there were no need to train neural networks from scratch. Our model achieves best validation accuracy of 60.37\% which is approximately 8.5\% higher, than VGAF dataset baseline and is competitive with existing works, audio and video modalities.
\end{abstract}

\hspace{-4.5mm}\textbf{CCS Concepts} \hfill \break
\textbullet \textbf{Computing methodologies$\sim$Artificial intelligence$\sim$
Computer vision}; \textbullet \textbf{Human-centered computing$\sim$Human computer interaction (HCI)}
\section{Introduction}
\label{sec:intro}
Emotion recognition is difficult and important task. Understanding emotions in groups of people is vital not only for every individual of a group, but also for people with different background and cultures. Moreover, knowledge about common emotion of group could be of interest for business, learning, healthcare, surveillance and robotics.

Early affective computing was focused on individuals, while in recent years more research is done for groups of people from raw, uncontrolled data (i.e. "in the wild") \cite{emotiw5}. Results performed on these type of data are easier adoptable to real-world situations such as surveillance cameras or videos from internet. 

Dataset used in training was presented in EmotiW 2020 Audio-Video Group Emotion Recognition grand challenge \cite{emotiw2020} and the exact challenge is Video level Group AFfect (VGAF). The task of this competition was to classify group emotion into three different categories: negative, neutral and positive. The biggest challenges of this dataset are: different lightning conditions, languages, video quality, frame rate, occlusions and intersection of people. One of the approach to handle such data, is to use only one modality \cite{singleaudio}, \cite{savchenkofacial}. Another option are two-stage models, where stages are feature extraction and modality fusion respectively \cite{berkleypaper}.

Despite the fact, that unimodal models show decent results, when given poor input information, this problem can not be compensated by another modality, which will hurt the performance. As for two-stage models, fixed feature extractors can not be fine-tuned for a specific task with respect to the information from low-level fusion layers.  

To address these issues we propose our model with following features: 
\begin{itemize}
	\item Model was trained fully end-to-end, compensating the problem of missing information about modalities interaction. Moreover, if one modality does not carry a lot of useful information, another one can mitigate this problem.
	\item We do not freeze layers throughout our model, which allows it to be fully optimized to a given task and helps model achieve solid results using models from different domains effectively.
\end{itemize}

\begin{figure*}[htp]
	\centering
	\includegraphics[width=\textwidth]{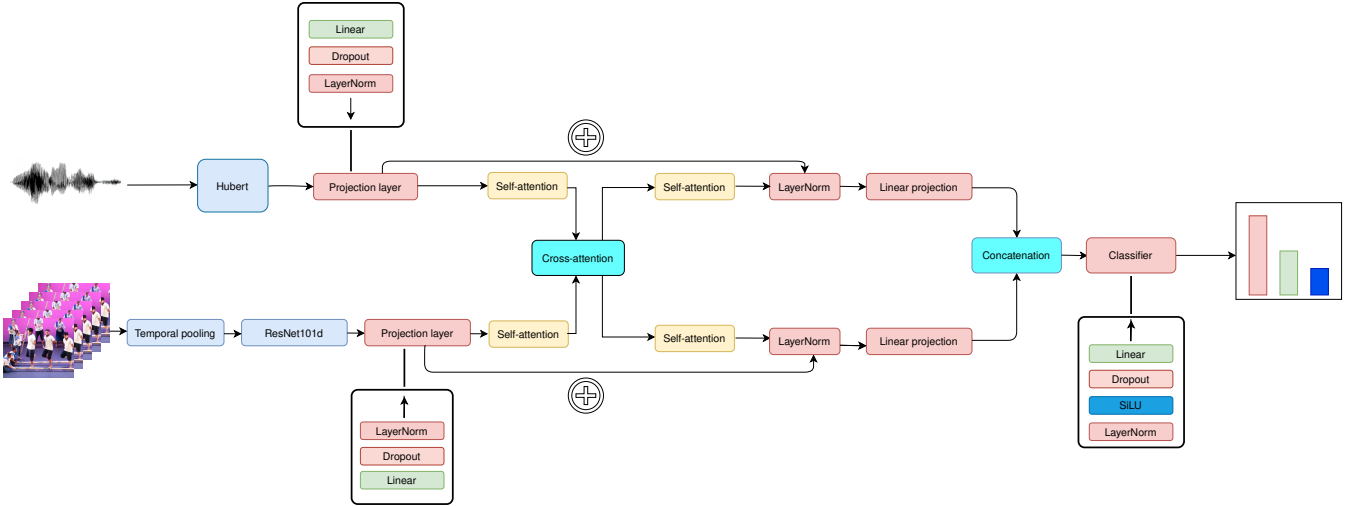}
	\caption{Proposed end-to-end architecture for group emotion recognition.}
\end{figure*}

\section{Related work}
Automatic human emotion recognition is a topic of active research for nearly a decade \cite{old_emo_rec},  \cite{multimodal_em_rec}, \cite{iemocap}. In early works on multimodal learning there were several techniques proposed, such as early fusion \cite{early_fusion} and late fusion \cite{late_fusion}. While the idea behind these techniques are simple, they show decent results and are still being widely adopted for multimodal tasks. Recently, researchers have been working not only on fusion techniques, but also on multimodal architectures, where pairs of different modalities' are being fed to network. These architectures, for instance, are Tensor Fusion Networks \cite{tensor_fusion}, LXMERT \cite{lxmert}, ClipBERT \cite{clipbert}, VATT \cite{vatt}, VILT \cite{vilt}. Due to increase in computational power in recent years, a lot of work was done using multimodal learning in different areas, such as question answering \cite{qa_complex}, image captioning \cite{image_capt},  emotion recognition \cite{endtoendemrec},  affect recognition \cite{affect_rec}. 

Previous results on VGAF \cite{emotiw2020} dataset were obtained using two-stage models, where at first features were extracted using fixed models and then, late fusion was used for fusion of extracted features \cite{berkleypaper}, \cite{sota_emotiw}, \cite{k_inj_emotiw}, \cite{audio_only_emotiw}, \cite{low_level_paper}, \cite{second_best_emotiw}, \cite{av_emotiw}. The best result for this dataset was shown by the winners of Audio-Visual Group Emotion Recognition challenge of EmotiW2020 \cite{sota_emotiw}, where team uses 5 different modalities and achieves 74.28\% on validation set.

\section{Dataset}
The VGAF dataset \cite{emotiw2020} contains 2661 video clips for training and 766 for validation. Dataset was acquired from YouTube with tags, such as "fight", "holiday" or "protest", videos from which characterize different emotions. Each video was cropped in clip of 5 seconds length. The data is contains 3 classes -- Positive, Neutral and Negative corresponding to the 3 emotion categories. Challenges of dataset are different contexts in every video, various resolutions, frame rates, multilingual audio, which is serious abstraction for the vast majority of available models. Moreover, as there are no labels of what exactly language is in the video, it makes it impossible to collect additional text transcription (modality) using automatic tools.

\section{Methodology}
\subsection{Encoders}
We describe the problem as follows. Let X = \begin{math}\{(v_i,a_i)\}_{i=1}^{I} \end{math} be a sample of data, where \textit{I} is a number of multimodal samples, \textit{$v_i$} is a sequence of RGB video frames and \textit{$a_i$} is a raw audio of a given sample.

First, we extract 8 equally distributed frames. Vision encoder accepts a sequence of extracted frames  \begin{math}\forall v_i\in \mathbf{R^{C\times{T}\times{H}\times{W}}}\end{math}, where \textit{C}, \textit{T}, \textit{H}, and \textit{W} are the number of channels, temporal size, height, and width, respectively. For the vision encoder we use approach, inspired by ClipBERT \cite{clipbert}, namely we applied mean-pooling (\,we will define this operation as \textit{M})\, to aggregate temporal information of a frames sequence, which is inexpensive way to make use of temporal information. We use pretrained 2D ResNet101d \cite{resnetd} for feature extraction. During our experiments we tried several backbone architectures described in \textbf{Table \ref{table:1}}. We decided not to use 3D CNNs, because it can highly increase time of training without advantage in accuracy \cite{TSM}, \cite{pseudo3d}. To pass extracted feature maps further, in attention layers, we flatten embedding on last two dimensions and proceed with passing vector to projection layer (defined as \textit{PL} shown on Fig.1), which projected embeddings of different modalities to a common space. During current research instead of projection layer, layer module was considered, where GeLU \cite{gelu} activation function and additional projection was added, but such a module lead to faster overfitting and shows approximately 3\% drop in acccuracy.

\begin{table}[ht]
\centering
\begin{tabular}{ |c|c| } 
    \hline
    Backbone & Accuracy \\ 
    \hline
    ResNet101d \cite{resnetd} & \textbf{60.37\%} \\ 
    \hline
    ResNet50d \cite{resnetd} & 58.44\% \\ 
    \hline
    RexNet100 \cite{rexnet} & 57.98\% \\ 
    \hline
    ResNet50 \cite{resnet}  & 56.83\% \\ 
    \hline
\end{tabular}
\caption{Various vision encoder backbones and overall model validation accuracy.}
\label{table:1}
\end{table}

We extracted raw audio data at sampling rate of 16000 Hz, and pass it to Hubert \cite{Hubert}.  We choose this model, because it was trained in self-supervised manner, which provides more robust representations and can handle multilingual data. To pass embeddings further, in self-attention layer, projection layer (shown on Fig.1) was used.

We define process of extracting embeddings of audio and video encoder stages as follows: \begin{center}\begin{math}\begin{array}{l}\forall v_i: v_{emb} = PL(ResNet101d(M(v_i))) \\
\forall a_i: a_{emb} = PL(Hubert(a_i)) \end{array}\end{math}\end{center}   for audio embeddings, where $v_{emb}, a_{emb}\in \mathbf{R^{S\times{N}}}$, \textit{S} is a sequence length and \textit{N} is a number of features.

\subsection{Attention layers}
In this section we will review attention layers used in our model: self-attention and cross-attention. The main purpose of these layers in our model is aligning multimodal information from encoding embedding after encoders.

\textbf{Self-attention} was initially intorduced in \cite{bahdanauattention} and described as layer for extracting information from a set of \textit{context} vectors to \textit{query} vector. Formally, it can be written as \\* $$Attention(Q, K, V) = softmax(\frac{QK^T}{\sqrt{d_k}})V$$ where \textit{K} and \textit{V} are context vectors and \textit{Q} is a query vector. For our model we used multi-head attention (MHA), which was introduced in \cite{attentionyouneed} and can be defined as
$$ MHA(Q, K, V) = Concat(head_1, .., head_h)W^O$$
$$\textrm{where}\; head_i = Attention(Q W^Q_i, K W^K_i, V W^V_i) $$
\begin{math}W^Q_i, W^K_i, W^V_i\end{math} are learnable parameters of query, key and value respectively. Unlike the BERT \cite{bert}, in our model self-attention is not used for text data, but for audio and visual embeddings. In the context of our model input to the self-attention is \begin{math}Q = K = V = (a_{emb} \oplus v_{emb})\end{math}. 

\textbf{Cross-modal attention} has similar definition, but instead of computing dot product between the same vector for query and key, it makes use of multimodal nature of video, i.e. there are two cross-modal attentions, one takes K = V = $a_{emb}$ and Q = $v_{emb}$ as input; another one takes K = V = $v_{emb}$ and Q = $a_{emb}$. Such attention enables one modality for receiving information from another modality and helps align them.

Following architecture decisions of \cite{attentionyouneed}, we have added \textbf{skip connections}, which add weights from layer before attention block with weights from layer after it. To prevent exploding gradient problem, Layer Normalization was applied before penultimate projection layers.
\section{Training}
We train our model using Adam optimizer \cite{Adam}, with learning rate $1e^{-4}$ and weight decay $1e^{-3}$. As encoder parts of our models were already pretrained and only being fine-tuned during training procedure layers of this part of the model was trained with lower learning rate multiplied by factor of $1e^{-2}$. 

One of the biggest challenge of learning multimodal neural networks is the fact, that they are exposed to severe overfitting. Usual regularization techniques are often ineffective \cite{regulariztion} for these networks. To mitigate this problem we use \textbf{label smoothing} with $\epsilon = 0.2$, which makes neural network be less "confident" about class it predicts.

\begin{table}[ht]
\centering
\begin{tabular}{ |c|c|c| } 
    \hline
    Models & Modality & Accuracy \\ 
    \hline
    K-injection network \cite{k_inj_emotiw} & A+V & 63.58\% \\ 
    \hline
    ResNet50 + BiLSTM \cite{av_emotiw} & A+V & 61.83\% \\ 
    \hline
    Hubert+ResNet101d (ours) & A+V & \textbf{60.37}\% \\ 
    \hline
    Inception + LSTM \cite{emotiw2020} (baseline) & A+V & 52.09\% \\ 
    \hline
\end{tabular}
\caption{Results on validation data for two modalities on VGAF dataset. A, V, F states for audio, video, face accordingly.}
\label{table:2}
\end{table}

\begin{figure}
	\includegraphics[width=78mm]{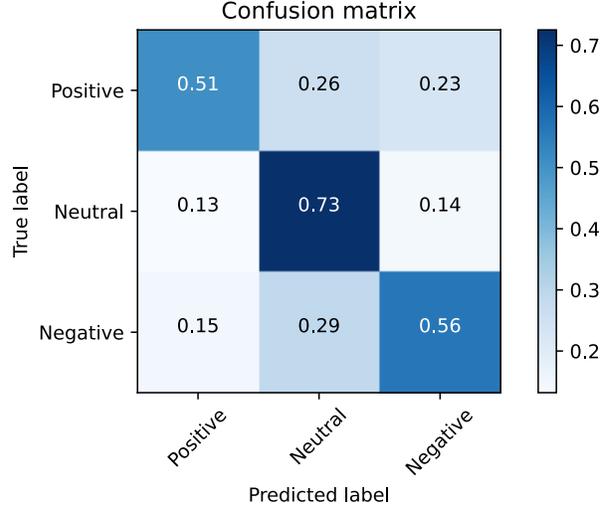}
	\caption{Confusion matrix for predictions of our model.}
\end{figure}

\section{Results}
We use accuracy as evaluation metric of our model. It achieves an overall accuracy of 60.37\% and  outperforms baseline by a margin of 8.5\%. In \textbf{Table \ref{table:2}} comparison of all available teams results, who used audio and video modalities for their final predictions, is shown. It can be seen, that our model has a competitive results compared to other works. The best model consists of ResNet101d and Hubert \cite{Hubert} as encoders for video and audio. The most challenging class for our model is "Positive", it can be explained by the challenging nature of the dataset and hard interpretation of emotions by themselves. Moreover, some videos in dataset have similar context, but different emotions and labels for them. Classifying "Neutral" videos as "Negative" ones can be a problem, which emerges from the fact, that there are big number of protests in dataset, where some of them are peaceful and others are aggressive.

\section{Conclusion}
Group video emotion recognition is a challenging task, especially for "in the wild" data. In this paper we present model for VGAF dataset from Audio-Visual Group Emotion Recognition challenge of EmotiW2020. Two novel approaches is used for our model.  Our model was trained end-to-end and optimized fully during training process, which help us achieve noticeable result of 60.37\% validation accuracy, which outperforms baseline significantly and can perform practically on par with existing bimodal audio-visual models.

\bibliographystyle{IEEEbib}
\bibliography{references} 

\end{document}